\documentclass[10pt,twocolumn,letterpaper]{article}

\usepackage{3dv}
\usepackage{times}
\usepackage{epsfig}
\usepackage{graphicx}
\usepackage{amsmath}
\usepackage{amssymb}
\usepackage{color}

\usepackage{bbm}
\usepackage{bm}
\graphicspath{{generated-images/}{images/}}
\DeclareGraphicsExtensions{.pdf,.jpeg,.png,.jpg}
\usepackage[caption=false,font=footnotesize]{subfig}

\usepackage[pagebackref=true,breaklinks=true,letterpaper=true,colorlinks,bookmarks=false]{hyperref}

\threedvfinalcopy %

\def\threedvPaperID{200} %

\newcommand{\vx}{\vec{x}}
\newcommand{\patchx}{\mathcal{P}(\vx)}
\newcommand{\inlx}{\mathcal{I}(\vx)}
\newcommand{\meth}{\mathcal{M}}
\newcommand{\img}{I}
\newcommand{\scr}{S}
\newcommand{\coords}{C}
\newcommand{\loss}{\mathcal{L}}

\newcommand{\olap}{o}

\newcommand{\reffig}[1]{Figure~\ref{#1}}
\newcommand{\refsec}[1]{Section~\ref{#1}}
\newcommand{\qmarks}[1]{``#1"}
\newcommand{\pdv}[1]{\frac{\partial}{\partial #1}}

\definecolor{somegray}{rgb}{0.5, 0.5, 0.5}
\newcommand{\darkgrayed}[1]{\textcolor{somegray}{#1}}
\makeatletter
\newcommand*\titleheader[1]{\gdef\@titleheader{#1}}
\AtBeginDocument{%
  \let\st@red@title\@title
  \def\@title{%
    \vskip-3em
    \bgroup\normalfont\large\centering\@titleheader\par\egroup
    \vskip1.5em\st@red@title}
}
\makeatother

\titleheader{\darkgrayed{This paper has been accepted for publication at the\\
IEEE International Conference on 3D Vision (3DV), Qu\'{e}bec City, 2019.
\copyright IEEE}}

\ifthreedvfinal\fi

\title{SIPs: Succinct Interest Points \\ from Unsupervised Inlierness Probability Learning}

\begin{document}

\author{Titus Cieslewski\\
University of Zurich
\and
Konstantinos G. Derpanis\\
Ryerson University and\\
Samsung AI Centre Toronto
\and
Davide Scaramuzza\\
University of Zurich
}

\maketitle

\begin{abstract}
A wide range of computer vision algorithms rely on identifying sparse interest points in images and establishing correspondences between them.
However, only a subset of the initially identified interest points results in true correspondences (inliers).
In this paper, we seek a detector that finds the minimum number of points that are likely to result in an application-dependent \qmarks{sufficient} number of inliers $k$.
To quantify this goal, we introduce the \qmarks{$k$-succinctness} metric.
Extracting a minimum number of interest points is attractive for many applications, because it can reduce computational load, memory, and data transmission.
Alongside succinctness, we introduce an unsupervised training methodology for interest point detectors that is based on predicting the probability of a given pixel being an inlier.
In comparison to previous learned detectors, our method requires the least amount of data pre-processing.
Our detector and other state-of-the-art detectors are extensively evaluated with respect to succinctness on popular public datasets covering both indoor and outdoor scenes, and both wide and narrow baselines.
In certain cases, our detector is able to obtain an equivalent amount of inliers with as little as $60\%$ of the amount of points of other detectors.
The code and trained networks are provided at \url{https://github.com/uzh-rpg/sips2_open}.
\end{abstract}

\section{Introduction}

A wide range of computer vision applications rely on establishing point correspondences between images.
These correspondences can be established densely for every pixel \cite{Rocco18nips}, but it is often of interest to instead establish only a sparse set of correspondences.
A sparse set makes many algorithms, such as visual odometry or bundle adjustment, far more tractable, both in terms of computation and memory.
Furthermore, in multi-agent scenarios, using a sparse set of points to represent an agent's observation means that less data needs to be exchanged.
\begin{figure}
  \centering
  \includegraphics[width=\columnwidth]{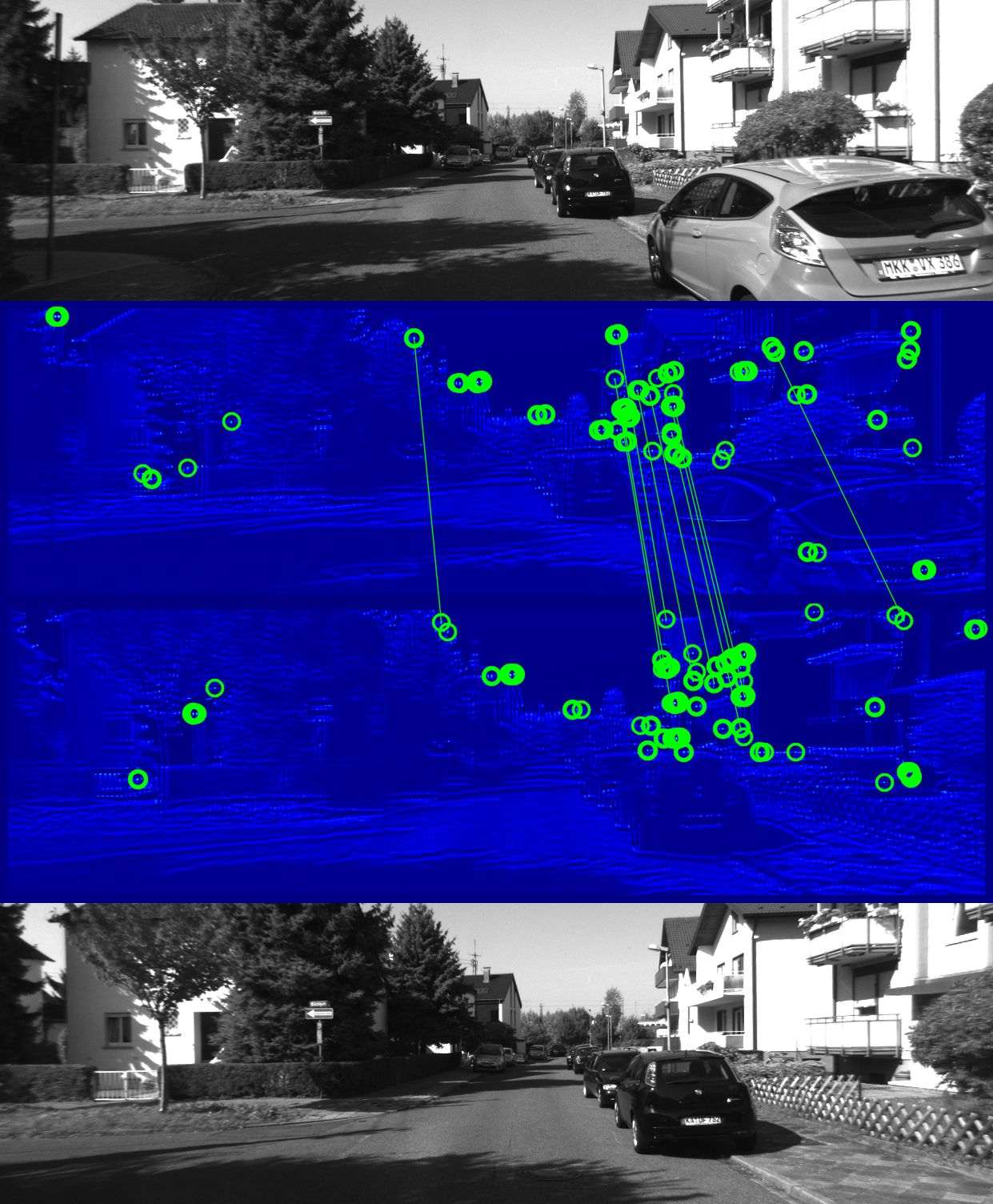}
  \caption{Not many inlier correspondences (green lines) are necessary to establish a good relative pose estimate between two image frames.
  In this paper, we look for a \emph{succinct} interest point detector -- a detector that results in sufficient inliers after extracting as little interest points (green circles) as possible.
  We introduce a corresponding metric, and a novel, unsupervised way to train interest point detectors.
  In blue, the score map of our detector.
  Local maxima may not be visible in print, see \reffig{fig:localized} for a close-up.}
  \label{fig:eyecat}
\end{figure}

Typically, a set of interest points is first identified with a detector.
However, only a fraction of these points will result in correct correspondences, as some of them will fail to be matched to points in the other image.
Of the points that do match, some will further fail geometrical consistency, resulting in an even smaller set of inlier points.
In this work, we seek to find the minimal set of points to extract in order to obtain a specific number of inliers{, as shown in \reffig{fig:eyecat} for $10$ inliers}.
While approaches exist to reduce an already given set of detected points \cite{Turcot09iccvw, Yi18cvpr}, we focus instead on directly detecting a minimal set.

To quantify this objective, we introduce the {\it $k$-succinctness} metric that builds on top of the widely used {\it matching score} \cite{Mikolajczyk05ijcv}.
The matching score indicates what fraction of a given set of points results in inliers, but it depends on the number of points extracted.
In contrast, the $k$-succinctness metric answers the question of how many points need to be extracted with a given detector, and in the context of a given point matching algorithm, to yield a certain number $k$ of inliers.
The scalar $k$ is a parameter of the metric, and depends on the application at hand.
In our experiments, we consider P3P pose estimation \cite{Gao03pami} as an application.
Given three 3D reference points in the object frame and their corresponding 2D projections, P3P determines four possible solutions for the orientation and position (pose) of a calibrated perspective camera \cite{Hartley03book}.
A fourth point is needed to disambiguate between the four solutions.
So theoretically, only four inliers are sufficient to establish a pose estimate, but we show experimentally that instead, choosing $k>4$ results in significantly better pose estimation performance.

Alongside the succinctness metric, we propose a new, simple training methodology for interest point detectors.
The network is simply trained to predict, for every pixel, the probability of resulting in an inlier, after being subjected to non-maxima suppression and descriptor matching.
Furthermore, we show for the first time how an interest point detector can be trained without the necessity of provided correspondence labels or pre-calculated depths, but merely with KLT tracking.
The proposed detector and other state-of-the-art detectors are evaluated with respect to succinctness on three datasets representing both indoor and outdoor scenes, and image pairs with narrow and wide baseline.

To summarize, our contributions are as follows:
\begin{itemize}
\setlength\itemsep{-.5em}
\item The introduction of the {\it succinctness} metric, which evaluates descriptors by the minimum number of points that need to be extracted to achieve a certain number of inliers.
\item A novel training methodology for interest point detectors that learns to predict the probability of a pixel being an inlier correspondence, in a self-consistent manner.
The proposed method requires the least amount of training data pre-processing to date.
\item An extensive evaluation of the proposed and state-of-the-art detectors with respect to succinctness on datasets covering both indoor and outdoor scenes, and wide and narrow baselines.
\end{itemize}

\section{Related work}

{\bf Traditional feature detectors}
Classically, interest points have been selected from distinctive image locations, that is, image locations that significantly differ from neighboring image locations.
Whether an image location is distinctive can be determined explicitly \cite{Moravec80thesis} or using a first-order approximation, such as the Harris~\cite{Harris88} or Shi-Tomasi~\cite{Shi94cvpr} detector.
Alternatively, distinctive interest points can be detected by convolving the image with a suitable kernel, such as the Laplacian of Gaussian (LoG), or its faster approximation, the Difference of Gaussians (DoG) kernel, prominently used in SIFT~\cite{Lowe04ijcv}.
Other filter-based detectors have been proposed by \cite{Mainali13ijcv,Bay08cviu}.
Finally, another method for choosing interest points is to identify image regions that explicitly resemble a sub-type of distinctive points, such as corners.
Efficient ways to do this have been proposed by \cite{Guiducci88prl,Smith97ijcv,Trajkovic1998ivc} and have found widespread popularity with \cite{Rosten06eccv} due to its highly efficient implementation.
An interesting challenge in interest point detection is ensuring that the detection is independent of scale and affine transformations.
Scale invariance can be achieved using multi-scale detection \cite{Mikolajczyk04ijcv}, while affine invariance, in particular invariance to rotation is already given for most aforementioned detectors.

The problem with relying on distinctiveness as the interest point selection criterion is that it does not necessarily result in high repeatability, unless all distinctive points are selected.
A more refined interest point selection criterion is needed if one wants to preserve repeatability when extracting less points.
Some of the aforementioned works have attempted to derive such a criterion based on models or heuristics.
An alternative and more promising method to deriving this criterion is data-driven, using machine learning.
An early method based on a neural network has been proposed in \cite{Dias95nnc}, though it was limited to three layers due to the computational constraints of the time, and also only applied to the edge regions of an image.
Subsequently, \cite{Richardson13icra,Trujillo06icpr,Verdie15cvpr} used other types of regression functions to learn detectors.

{\bf CNN feature detectors}
With the recent popularity of convolutional neural networks (CNNs), CNNs have also been considered for interest point detection.
They are particularly well-suited for this, as a per-pixel interest score can be directly calculated with a series of convolutional layers.
LIFT~\cite{Yi16eccv} was the first to exploit CNNs for several components of the point correspondence process.
It uses a separate network for detection, orientation estimation and description of interest points.
This work is trained on patches provided by a SfM algorithm executed a priori.
{Lenc and Vedaldi}~\cite{Lenc16eccvw} train an interest point detector using a \qmarks{covariance constraint}, which trains the response of a network to be invariant to viewpoint changes.
{Savinov et al.}~\cite{Savinov17cvpr} generalize this constraint to enforce consistent ranking of responses of corresponding points between viewpoints.
Superpoint \cite{Detone18cvprw} proposes a sophisticated training method that involves first fitting a network to detect labeled corners in a synthetic dataset, and later to train invariance on real images by warping real images.
In contrast to previous work, detector and descriptor are trained jointly.
Similarly, LF-Net~\cite{Ono18nips} is trained on the full point correspondence algorithm.
The loss acting on the LF-Net detector is similar to the covariance contraint of \cite{Lenc16eccvw}, augmented with a heuristic to favor responses that resemble narrow Gaussian kernels.
Additionally, a loss requiring point descriptors to match also acts on the detector.
The idea to explicitly favor sharp responses has also been presented in \cite{Zhang18cvpr}.
We find that in our approach, peakedness happens without the need for a specific loss, see \refsec{sec:nms} and \reffig{fig:localized}.
Prior work relied on synthetic images, synthetic warping, ground truth homopgraphies or depth for their training.
LF-Net~\cite{Ono18nips}, which has appeared in parallel to this work, uses COLMAP~\cite{Schoenberger2016cvpr} to obtain depth estimates from potentially uncalibrated image sequences.
In contrast, we show how to train our network using only KLT tracking performed on uncalibrated image sequences, requiring less training data pre-processing.

{\bf Interest point detector evaluation}
The traditional way to evaluate interest point detectors is repeatability, as first introduced in \cite{Mikolajczyk02eccv}.
Repeatability quantifies the overlap of, generally, detected affine regions in one image and the ground truth correspondence of the affine regions detected in the other image.
Specifically, if a detector returns only points, the affine regions can be approximated as circles that are either uniform or scale-dependent.
Repeatability was used to benchmark traditional detectors in \cite{Mikolajczyk05ijcv, Mikolajczyk05pami}.
More recently, \cite{Lenc18bmvc, Komorowski18eccvw} have used repeatability to benchmark more modern detectors.
Repeatability, however, has two issues as a metric.
Firstly, performance under repeatability depends on the number of extracted points.
In the most extreme case, when every point is \qmarks{extracted as interest point}, repeatability technically reaches its maximum value.
{Lenc and Vedaldi}~\cite{Lenc18bmvc} address this issue to some extent by evaluating repeatability at different specific point counts.
Secondly, repeatability does not take descriptor and matching performance into account.
Consider a detector that detects all corners of a large checkerboard.
While this detector will be very repeatable, it is likely that its points will be useless, as typical feature matchers will not be able to distinguish them.
The latter problem is addressed with the matching score \cite{Mikolajczyk05ijcv}.
Unlike in repeatability, affine region overlap is only considered if the corresponding interest points have also been matched.
However, also the matching score performance depends on the point count.
In contrast, succinctness does not depend on the point count.
Rather, it reflects the point count that is necessary to obtain $k$ inliers.
While this shifts to a dependence on $k$, $k$ should typically be given by the application.

{\bf A posteriori compression}
An alternative approach to finding minimal sets of interest points would be to first extract and describe a large set of interest points, and then reduce them to a minimal set sufficient for localization in post-processing.
Several approaches exist where this reduction is based on whether specific features are consistently detected in multiple images observing the same scene \cite{Turcot09iccvw,Li10eccv,Park13cvprw,Cao14cvpr,Dymczyk15icra}.
Learning to predict from single images whether already-extracted features are likely to be matched has also been proposed \cite{Hartmann14cvpr,Dymczyk16threedv,Yi18cvpr}, but again as a filtering step of already-extracted features.
Instead, we directly identify succinct interest points in the detector.

\section{Succinctness}\label{sec:succ}

Succinctness answers the question: {\it how many interest points $n_k$ need to be extracted by a detector to achieve $k$ inlier correspondences?}
Hence, succinctness will not only depend on the detector, but also the descriptor, matching, potentially other parts of a feature matching pipeline, and whatever method is used to distinguish inlier correspondences from outlier correspondences.
Specifically, while this distinction can be made based on ground truth information, we also allow for it to be achieved using another method, such as geometric verification using P3P~\cite{Gao03pami} with RANSAC~\cite{Fischler81cacm}.
Furthermore, just as for other metrics, succinctness will depend on the data on which it is evaluated.
In particular, we need a mechanisms to aggregate $n_k$ across multiple image pairs, since $n_k$ will be different for every evaluated image pair.
For this, we could use histograms, but histograms introduce a dependency on the selected bins.
Instead, we propose to use a cumulative aggregation.
Specifically, we propose the {\it succinctness curve}:
the $x$-axis represents a specific amount of extracted points $n$, and the value on the $y$ axis is $s(n)$, the fraction of pairs whose $n_k$ is lower than $n$.
$s(n)$ can also be thought of as the fraction of image pairs that achieve at least $k$ inliers if $n$ points are extracted -- we assume that inlier count increases monotonically with the amount of extracted points.
An example of such curves can be seen in \reffig{fig:succ_kt}.

To quantify succinctness over a whole dataset in a single number, we use the area under this curve, up to a specific $n_\text{max}$.
Formally:
\begin{equation}
\text{AUC-}n_{\text{max}} = \frac{1}{n_{\text{max}}} \int_{n=0}^{n_{\text{max}}} s(n) dn \label{eq:auc} .
\end{equation}
This number assumes values between $0$ and $1$.
For example, if $\text{AUC-}n_{\text{max}} = 0.8$, this means that either ($80\%$ of image pairs obtained $k$ inliers with $0$ extracted points \footnote{This is of course an impossible example, but serves for illustration.} and $20\%$ failed to obtain $k$ inliers with $n_\text{max}$ interest points) or (all image pairs obtained $k$ inliers with $0.2 \cdot n_\text{max}$ interest points), or something in between these two extremes.

To rapidly determine $n_k$ for a given image pair, we perform only one detection step in which we extract $n_\text{max}$ interest points in both images.
We then perform binary search, where the features of both images are cropped to the $n$ features with the highest score, and the rest of the feature matching system is executed on those features, until we find the smallest $n$ that achieves at least $k$ inliers.

\section{System Overview}\label{sec:meth}

\begin{figure}
  \centering
    \includegraphics[width=\columnwidth]{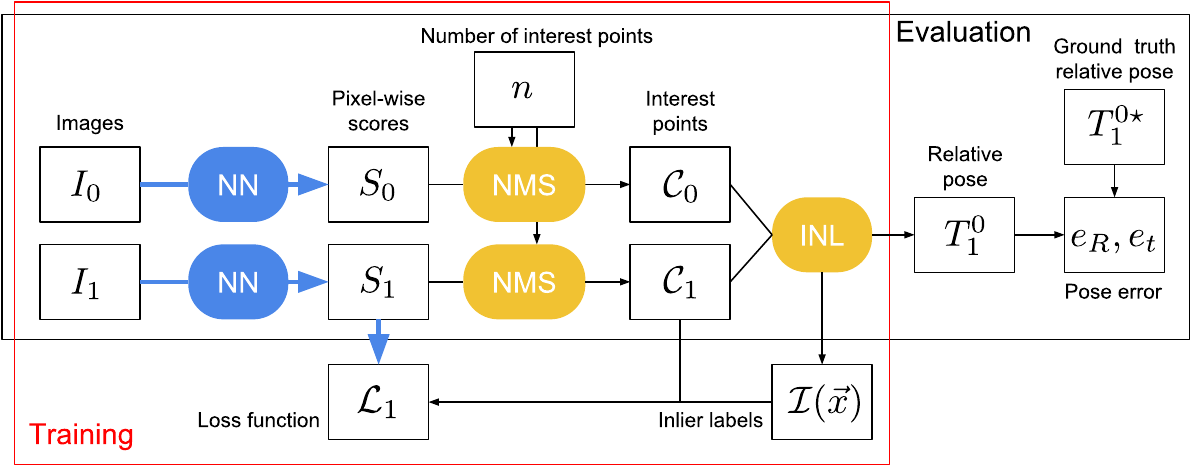}    
  \caption{The system we propose for training and evaluation.
  The neural network (NN) calculates a pixel-wise score map $S_i$ for each image $I_i$.
  The $n$ highest scored interest points $\coords_i = \{\vx\}$ are selected, using non-maxima suppression (NMS).
  During training, an inlier determination module (INL) matches the points and determines which matches are inliers $\inlx$.
  This information, together with interest point locations is used to formulate our loss $\loss$ on patches centered at the interest point locations, which is back-propagated during training (blue arrows).
  During testing and deployment, the inlier determination module is replaced with a pose estimator that uses P3P and RANSAC to find a relative pose $T_1^0$ that can be compared to ground truth for accuracy evaluation.
  }
  \label{fig:sys}
\end{figure}
Similar to recent learned interest point detectors, we use a fully convolutional neural network (\refsec{sec:arch}) to predict a per-pixel score map $\scr_i$ from an image $\img_i$.
\reffig{fig:sys} illustrates our system used for interest point detection, relative pose estimation, and network training.
Non-maxima suppression with a radius of $5px$ is used to find the $n$ highest-scoring points $\coords = \{\vx\}$ in $\scr$.
This constitutes the forward pass of the system for a single image.
For training, an inlier determination module (\refsec{sec:inloutl}) processes the interest points of two images and determines which of them are inliers.
The loss (\refsec{sec:loss}) is sparsely applied at the interest point locations.
The inlier determination module extracts and matches SURF descriptors \cite{Bay08cviu} and uses KLT tracking \cite{Lucas81ijcai} in an image sequence to determine inliers.
The choice of SURF as descriptor is somewhat arbitrary -- as we do not provide our own descriptor, we need to use an existing one, and SURF proved both practical to adopt and sufficiently well-performing.
The presented method can easily be used with any other descriptor.
Inlier determination is replaced with P3P~\cite{Gao03pami} with RANSAC~\cite{Fischler81cacm} in the succinctness and pose estimation evaluation as illustrated on the right side of \reffig{fig:sys}.
In RANSAC, we use the maximum inlier set after a fixed amount of $3000$ iterations.
This corresponds to finding the true solution among $10\%$ inliers with a probability of $95\%$.

\section{Training Methodology}

We want the selected points to have the highest probability of being inliers.
To that end, we train our neural network to predict this probability.
Selecting the points with highest scores thus implies selecting the points that will most likely result in inlier correspondences.

\subsection{Inlierness Probability Model}

We model point $\vx$ being an inlier $\inlx$ as a stochastic process conditioned on the patch around $\vx$, $\patchx$.
In our model, this process results in $\inlx$ with probability
\begin{equation}
P = P(\inlx | \patchx),
\end{equation}
and in $\neg\inlx$ with probability $1-P$.
This is an approximation, because $P(\inlx)$ of course depends on more than just $\patchx$, but, as we will show empirically, this approximation is accurate in many scenarios, and even transfers well between different environments.

Among the unmodeled conditions, there are ones that are not under our control, such as the relative pose between the cameras, and ones that are under our control, such as the interest point selection and matching method $\meth$.
For the relative pose between cameras, we assume that the two cameras point at the same scene while providing many different examples of relative poses during training to train the network in a viewpoint-invariant way.
As for the method $\meth$, we always use the method presented in \refsec{sec:meth}, and so $P$ is implicitly conditioned on $\meth$.
Note that this introduces a circular dependency, as the interest point selection itself depends on the predicition of $P$.
This means that there is no \qmarks{ground truth} for $P$ -- our system will start in an arbitrary state and if trained well, it will converge to a self-consistent state.

\subsection{Loss function}\label{sec:loss}

$P$ can be learned using the cross-entropy loss:
\begin{equation}\label{eq:cel}
\ell_{\inlx}(P) = -\log(P), \quad \ell_{\neg\inlx}(P) = -\log(1-P),
\end{equation}
where $\ell_{\inlx}(P)$ is the loss applied if $\inlx$ and $\ell_{\neg\inlx}(P)$ the loss applied otherwise.
To provide the reader with an intuition of why this is the case, we briefly re-iterate the corresponding theory.
In principle, given $\vx$, the loss should move $P$ up if it is under-estimated, suppress it if it is over-estimated, and maintain it if it is correct.
However, according to our model, during training only random samples $\inlx$ and $\neg\inlx$ will be provided.
Hence, the best thing that we can do is to make sure that this condition holds over a large quantity $m$ of training steps involving a particular $\vx$.
In $m$ such training steps, the expected amount of steps where $\inlx$ holds is $Pm$ and the expected amount of steps where $\neg\inlx$ holds is $(1-P)m$.
Thus, for these contradicting training steps to cancel each other out at the true $P$ in the long term, the gradients of the loss should satisfy:
\begin{equation}\label{eq:balance}
P\pdv{P}\ell_{\inlx}(P) + (1-P)\pdv{P}\ell_{\neg\inlx}(P) = 0.
\end{equation}
Similarly, the first term of \eqref{eq:balance} should be larger than the second one if $P$ is under-estimated and smaller if $P$ is over-estimated.
These conditions are satisfied by the cross-entropy loss \eqref{eq:cel}.

\subsection{Unsupervised inlier determination}\label{sec:inloutl}

Previous methods used for training interest point detectors either involved labeled data \cite{Detone18cvprw}, synthetic warping of planar scenes \cite{Detone18cvprw, Lenc16eccvw}, or 3D scenes annotated with ground truth depth \cite{Savinov17cvpr, Ono18nips}, or ground truth correspondence labels provided in another manner.
LF-Net~\cite{Ono18nips}, which has appeared in parallel to this work, uses COLMAP~\cite{Schoenberger2016cvpr} to provide dense depth estimates, which allows training from uncalibrated image sequences just as our method.
Instead, we propose to use Lucas-Kanade tracking (KLT) \cite{Lucas81ijcai} to determine correspondences.
This has the benefit of lower computation and storage requirements.

Given a pair of forward passes from two images, and the intermediate images in the sequence, the interest points are tracked image-to-image{, through all intermediate images,} using KLT.
{ This can be applied to wide baseline image pairs, provided the interest points can be tracked through the intermeidate frames.}
To discard poorly tracked points, each image-to-image tracking is verified with a bidirectional check, where a track is only considered successful if tracking forwards and backwards results in an error that is below one pixel.
Using just this information would allow us to train a repeatable detector, but we also want to take the used descriptor in consideration during training.
Hence, an interest point is only considered an inlier if the correspondence obtained from KLT is within 3pixels of the one obtained from descriptor matching.

Of course, this correspondence method could potentially be applied in conjunction with other losses, such as \cite{Ono18nips}. 
Similarly, our training method is not restricted to using this method for inlier determination.
For example, for calibrated stereo image sequences, we can find the depths of the interest points in one of the two images (of the training pair, not the stereo pair) using epipolar stereo matching.
Then, the depths can be used to perform P3P RANSAC with the interest points of the other images, resulting in inlier labels from P3P RANSAC, a method which even more closely mirrors our evaluation.
Alternatively to these methods, any method, or labels, providing ground truth correspondence can be used in a similar fashion to inlier determination using KLT tracking.
{ In particular, note that KLT tracking is not suitable to label correspondences when intending to train illumination invariance.
Illumination invariance can be trained using images without changes in viewpoint~\cite{Lenc18bmvc} or images where true pose and depths are known~\cite{Aanaes12ijcv}.}

\subsection{Sparse loss application}\label{sec:sprop}

Similar to recent work, we only apply our loss at sparse locations.
Specifically, the loss is only applied at the extracted interest points.
During training, $500$ interest points are extracted.
Ono et al.~\cite{Ono18nips} argue that a loss should be applied at all correspondences to reach convergence, but we find that this is not the case in our training framework.
The fact that we do not also need to train a descriptor, but use an already existing one, might facilitate convergence in our case.

As a consequence of the application of a sparse loss, it is possible that the probability prediction is only properly trained for the higher probability values, which are the values at which interest points are extracted.
Since we are only interested in inlierness probability prediction for interest points, however, this is not considered an issue.

\subsection{Avoiding degenerate relative poses}\label{sec:nms}
\begin{figure}
    \centering
    \includegraphics[width=.8\columnwidth]{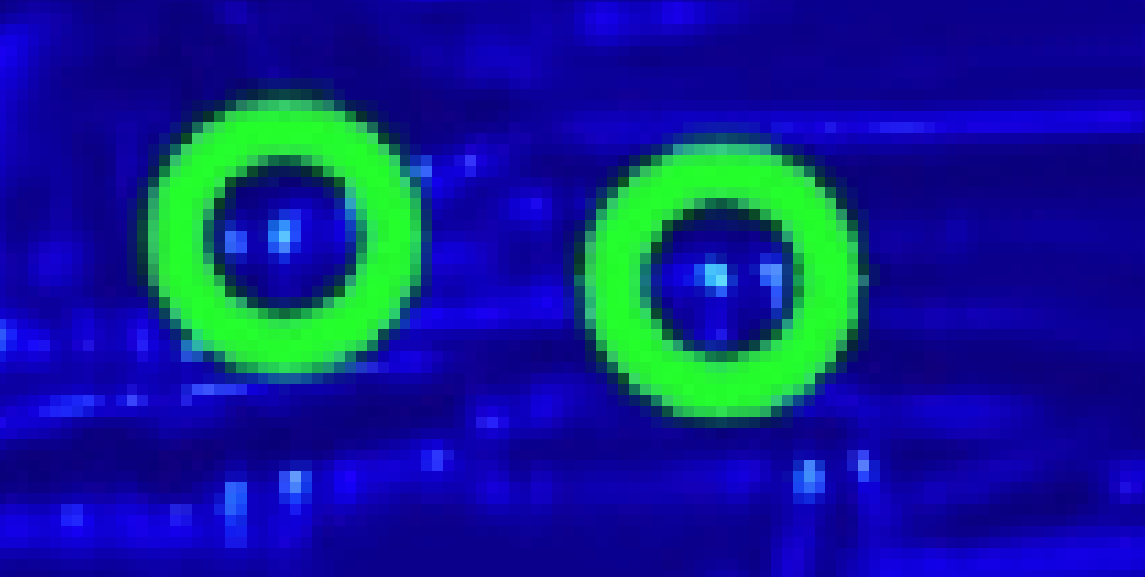}
    \caption{Close-up of our CNN response.
    Green circles indicate interest points selected in the image.}
    \label{fig:localized}
    \vspace{-3mm}
\end{figure}

To achieve good performance in pose estimation, features need to be well-distributed in the image.
There are generally two approaches to ensure such a distribution.
Firstly, non-maxima suppression (NMS), which ensures that there is a minimum distance between two selected interest points.
Secondly, binning, where a grid is overlayed over the image, and a specific amount of interest points is extracted in each bin.
While binning has proven to be useful in several applications \cite{Forster17troSVO, MurArtal17tro}, it seems to be at odds with the idea of succinctness, where repeatability at low point counts is key:
if a certain point crosses a bin boundary between the two images to be matched, it is very likely that it will only be selected in one of the images, as it potentially has to compete for the highest score with completely different interest points between the bins.
Instead, NMS seems to be a less intrusive way to distribute the points in the image.
We use a NMS radius of $5px$.
While this seems to be a relatively low radius, we find that applying it during training results in extremely peaked responses, without any further response peakedness loss as in \cite{Ono18nips, Zhang18cvpr}.
As can be seen in \reffig{fig:localized}, a local maximum is typically contained within $3 \times 3$ pixels, often with a single pixel having the clearly highest response.
Plenty of other examples can be seen by zooming into \reffig{fig:eyecat} and renderings provided in the supplementary material.

\subsection{Image pair selection}\label{sec:kltsel}

While some training datasets like HPatches \cite{Balntas17cvpr} provide pre-selected image pairs, we are also interested in training from uncalibrated image sequences.
Here, two images need to be from the same scene for training and evaluation to be meaningful.
This poses a problem, as we want to avoid manual annotation or additional labels such as ground truth pose.
LF-Net~\cite{Ono18nips} addresses the problem of pair selection in sequences by sampling pairs at specific time differences.
However, this would not guarantee scene overlap between the two images on general image sequences.
We instead determine scene overlap using KLT.
For every sequence, a preprocessing step is executed where points are sampled densely in the full image, at some distance to each other, to remain tractable.
These points are then tracked using KLT, and new points are detected to keep the image densely sampled.
Then, based on the tracks, we track for each image the fraction of tracked points in subsequent images (\qmarks{overlap}), until no points from the original image are tracked.
This information is stored and used during random pair selection to select two images such that the second image has a minimum overlap $\olap$ with the first image.

\subsection{Neural Network Architecture}\label{sec:arch}

Our network is a fully convolutional network with $3 \times 3$ kernels and unit strides throughout, and leaky ReLU activations, except at the final layer.
The final layer activation is a sigmoid activation to constrain the output to $]0, 1[$.
The only two hyperparameters we consider are the depth (layer count) and width of the network, $d$ and $w$.
We use $\frac{w}{2}$ channels in the first half layers and $w$ in the second half, except for the final channel which outputs a single channel.
We use two models for evaluation:
firstly, a multi-scale model that calculates score maps at five different scales and uses $12$ layers and a width of $w=256$.
The scale factor is $\sqrt{2}$ and all scales share the same weights.
Secondly, a single-scale model that uses $10$ layers and a width of $w=128$.

\section{Experiments}

Previous work~\cite{Komorowski18eccvw} has shown that learned interest point detectors tend to perform poorly on real-world datasets.
This motivates our first set of evaluations, which is on robotics datasets.
We then also evaluate on the wide-baseline HPatches dataset \cite{Lenc18bmvc}.
Unlike prior benchmarks, however, we evaluate succinctness and pose estimation accuracy.
Finally, we evaluate the accuracy of the inlierness probability prediction.

\subsection{Compared detectors}

We evaluate succinctness for SIFT \cite{Lowe04ijcv}, SURF \cite{Bay08cviu}, LF-Net~\cite{Ono18nips} (which has appeared in parallel to this work), SuperPoint \cite{Detone18cvprw}, and our proposed network.
For SIFT and SURF, we use the OpenCV implementations, while for LF-Net and Superpoint, we use the publicly available code and pre-trained weights.
Unfortunately, LFNet does not provide the scores of the detected points, preventing us to do the binary search for $n$ (\refsec{sec:succ}), so our comparison is based on extracting three specific interest point counts $n = (50, 100, 150)$.

\subsection{Succinctness on Robotics Datasets}

To evaluate succinctness on robotics datasets, we consider the outdoor/autonomous driving dataset KITTI \cite{Geiger13ijrr}, and the indoor/drone dataset EuRoC \cite{Burri15ijrr}.
While our system supports multi-scale extraction, experiments on a validation sequence of KITTI show that multi-scale does not improve performance on these rather narrow-baseline datasets, so we only train the system at a single scale.
For training the network used in these experiments, we only use image pairs from the TUM mono~\cite{Engel17pami} and Robotcars~\cite{Maddern17ijrr} datasets.
For testing, we use sequence {\tt 00} of the KITTI dataset and {\tt V1\_01} of EuRoC.

From each testing sequence, we randomly select $100$ image pairs according to the method outlined in \refsec{sec:kltsel}, with the minimum overlap $o$ set to $0.5$.
These pairs are consistent for all evaluated methods.
The indices of the image pairs and visualized samples are provided in the supplementary material.

The evaluation datasets come with ground truth poses, which allows us to evaluate pose estimation error as a function of the selected minimum inlier count $k$.
{We obtain relative pose estimates using P3P~\cite{Gao03pami} with RANSAC~\cite{Fischler81cacm}, where the 3D location of points of one of the images is obtained from stereo matching.
This can be done as both evaluation datasets were recorded with stereo cameras.}
Position and rotation errors are considered separately, but their distribution over the evaluation set is summarized as one number in a way similar to succinctness.
Consider a plot where the $x$ axis represents the error and the $y$ axis represents the fraction of image pairs that result in an error that is lower than $x$.
Then, we use the AUC-1 \eqref{eq:auc} (1 meter for translation and 1 degree for rotation) to represent the distribution.
For example, if the AUC-1 is $0.8$, then either ($80\%$ of the pairs have a perfect estimate and $20\%$ fail RANSAC) or (all pairs have an error of $0.2$), or something in-between these extremes.
For position error we use Euclidean distance in meters and for rotation error we use geodesic distance (angle of angle-axis representation) in degrees.

\begin{figure}
  \centering
  \includegraphics[width=\columnwidth]{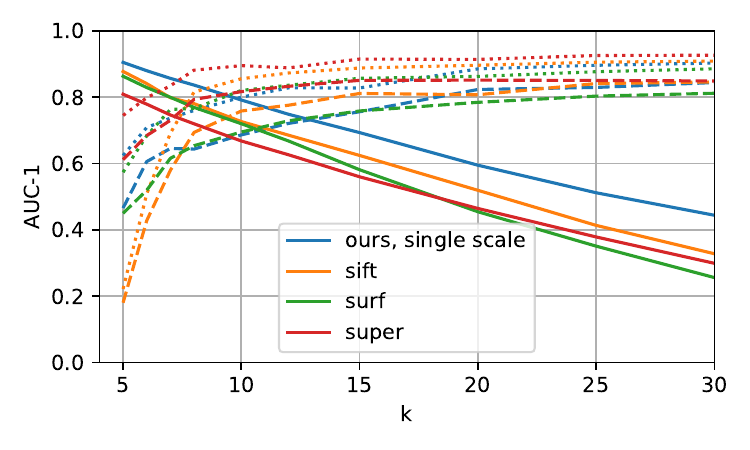} 
  \includegraphics[width=\columnwidth]{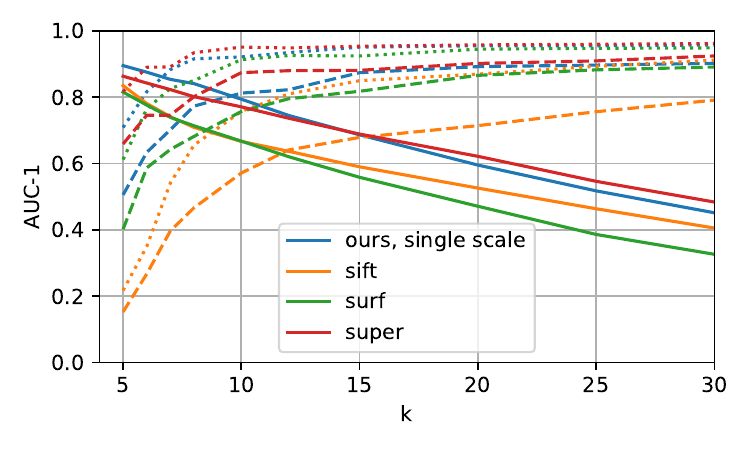}
  \caption{The role of the required inlier count in pose estimation quality, evaluated on KITTI {\tt 00} (top) and EuRoC {\tt V1\_01} (bottom, EuRoC uses AUC-$5$ for the rotation error).
  As we increase the required inlier count $k$, rotation (dashed) and translation (dotted) precision increases, but succinctness {(solid)} drops.}
  \label{fig:varyk}
  \vspace{-5mm}
\end{figure}

In \reffig{fig:varyk} we inspect how pose estimation quality changes as a function of required inlier count $k$, and what the corresponding succinctness is.
As we increase $k$ to $10$ in either dataset, significant improvements in pose accuracy can be observed.
Beyond $10$, accuracy further improves, but at a slower pace.
In terms of succinctness, out method outperforms baselines for all evaluated $k$ on KITTI and for $k<15$ on EuRoC, beyond which SuperPoint slightly outperforms our method.

\begin{figure}
  \centering
  \includegraphics[width=\columnwidth]{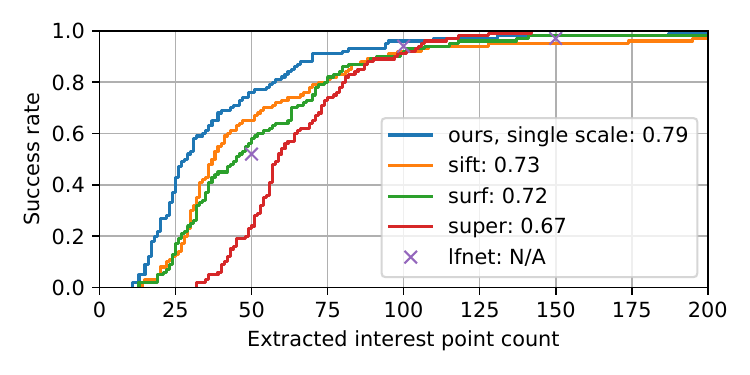}
  \includegraphics[width=\columnwidth]{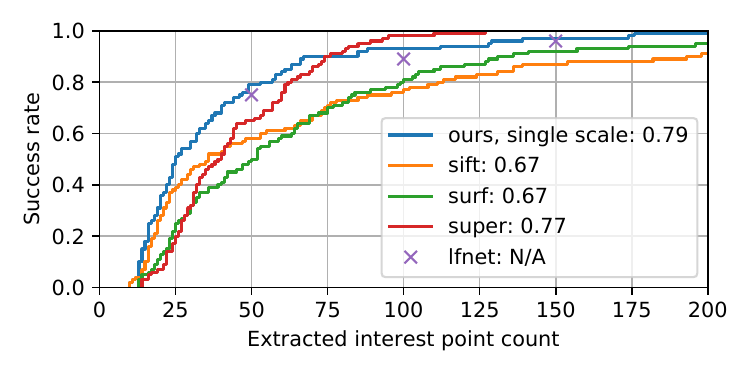}
  \caption{Succinctness and AUC-$200$ values on KITTI (top) and EuRoC (bottom) with $k=10$.
  At low point counts, the least amount of points need to be extracted with our detector.
  AUC could not be evaluated for LF-NET, as only $n  \in \{50, 100, 150\}$ were sampled.}
  \label{fig:succ_kt}
  \vspace{-3mm}
\end{figure}

In \reffig{fig:succ_kt}, we show succinctness curves for the specific value of $k=10$.
In contrast to \reffig{fig:varyk}, this allows us to see some more nuance, such as the fact that in the EuRoC dataset, our method has the highest success rate at $75$ interest points or less, but at higher interest point counts, SuperPoint has higher success rates on EuRoC.

\subsection{Pose accuracy with fixed point count}

\begin{figure}
  \centering
  \includegraphics[width=\columnwidth]{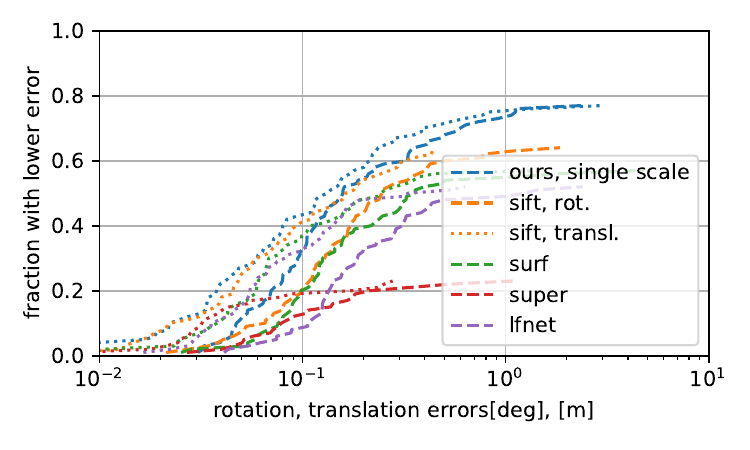}
  \includegraphics[width=\columnwidth]{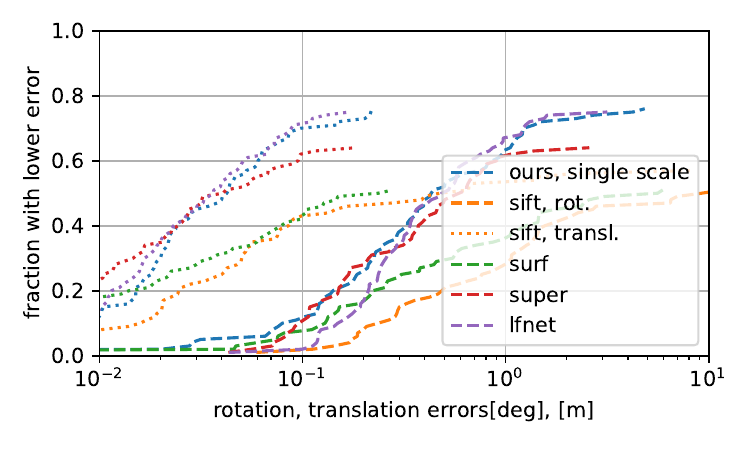}
  \caption{Pose estimation errors on KITTI (top) and EuRoC (bottom) when extracting 50 points and using the best inlier set with at least $10$ inliers after 3000 RANSAC iterations.
  {The dashed lines stand for rotation errors and the dotted lines for translation errors.}
  The curves end with the largest error obtained with $10$ inliers.}
  \label{fig:errs}
  \vspace{-5mm}
\end{figure}

Informed by these results, we consider a practical case, where the amount of extracted interest points is fixed to $50$, which, according to \reffig{fig:succ_kt}, should result in $80\%$ of testing samples having at least $10$ inliers.
\reffig{fig:errs} shows the distribution of rotation and translation errors of the relative pose estimates obtained when using the best inlier set with at least $10$ inliers after $3000$ RANSAC iterations.
As we can see, our method generally performs best at this low interest point number, with the other learned detectors performing similarly on EuRoC, but worse on KITTI.

\subsection{Inlierness probability prediction}

\begin{figure}
  \centering
  \includegraphics[width=\columnwidth]{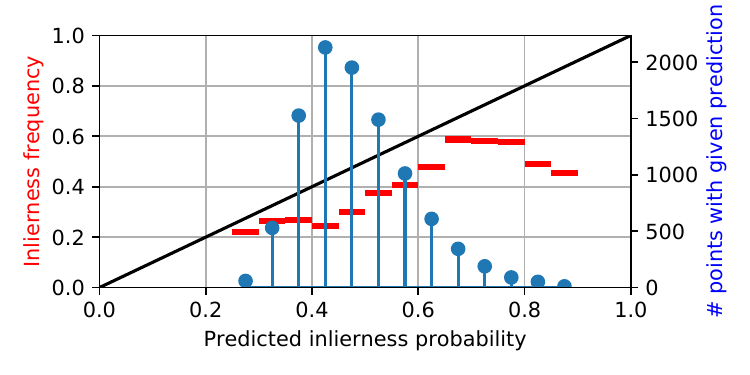}
  \includegraphics[width=\columnwidth]{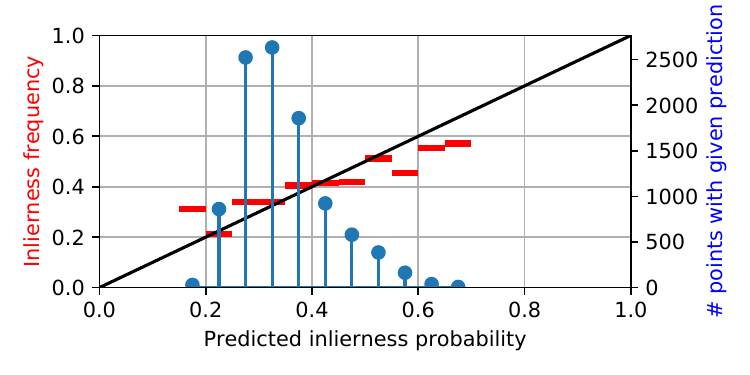} 
  \caption{
  Actual inlierness frequency versus predicted inlierness probability after extraction, matching, and P3P + RANSAC verification of $50$ interest points.
  Top: KITTI, bottom: EuRoC.
  Inlierness frequency is calculated within bins defined by the predicted probability, bin ranges are indicated by the width of the red lines, and their cardinality by the blue lines.
  The black line indicates perfect inlierness probability prediction.}
  \label{fig:valprob}
\end{figure}

We have trained our network to predict the probability of a pixel resulting in an inlier.
Besides resulting in a succinct interest point detector, \reffig{fig:valprob} shows that the ability to predict this probability generalizes well, even if tested with a smaller number of extracted points, here $50$ instead of $500$.
On KITTI, the predicted probability is somewhat overconfident, especially at the higher end.
However, only few points have a predicted probability beyond $75\%$.
Consequently, few samples were available to calculate the inlier frequency that the probability prediction is compared to.

\subsection{Succinctness on Wide Baseline}

\begin{figure}
  \centering
  \includegraphics[width=\columnwidth]{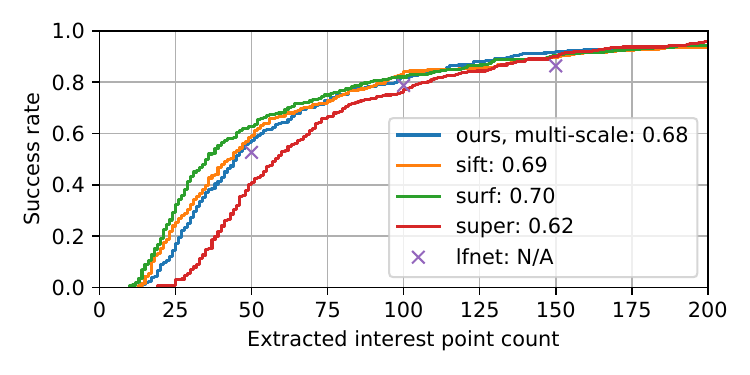}
  \caption{Succinctness and AUC-$200$ on HPatches with $k=10$.
  Wide baselines seem to be challenging for learned interest point detectors, in terms of succinctness.
  Among the learned detectors, our multi-scale detector trained on the HPatches training set performs best, almost as good as SIFT and SURF.
  AUC could not be evaluated for LF-NET, as only $n  \in \{50, 100, 150\}$ were sampled.}
  \label{fig:succ_hp}
  \vspace{-5mm}
\end{figure}

While the previous outdoor and indoor datasets represent scenes that are likely encountered in robotic applications, such as autonomous driving and autonomous drone flight, they exhibit a relatively narrow baseline.
To evaluate our method on more extreme viewpoint changes, we consider the viewpoint sets of HPatches \cite{Balntas17cvpr, Lenc18bmvc}, with the training and testing split as suggested by the authors.
For HPatches, we find that we need to train multi-scale, as several pairs in our validation sequence exhibit strong scale changes.
The network is trained on the HPatches training sets only.
For performance, we downscale the HPatches images such that the largest dimension is at most $1000px$, and convert the images to grayscale.
The results are shown in \reffig{fig:succ_hp}.
Consistent with the observation of \cite{Lenc18bmvc}, SIFT and SURF do very well here.
When trained on wide baselines, our network can achieve similar performance, as does LF-NET.
Here, Superpoint performs worst, but remarkably well considering that multi-scale is not explicitly expressed in its architecture.

\section{Conclusion}

In this work, we set out to find a detector that has good performance at low numbers of extracted interest points.
Extracting a minimum number of interest points is attractive for many applications, because it can reduce computational load, memory, and data transmission.
To quantify this goal, we have introduced the \emph{succinctness} metric, which measures the miminum required number of interest points to achieve a certain number $k$ of inliers.
In our experiments, the choice of $k$ has been driven by relative pose estimation using P3P and RANSAC.
We have evaluated several state-of-the-art detectors, along with a novel detector which can be trained with the least amount of training data labeling and pre-processing to date.
Our detector consistently exhibits top performance in terms of succinctness.
Additionally, it boasts an interpretable score which can be used to predict the probability that an interest point will yield an inlier.

\section*{Acknowledgments}
{This work was supported by the National Centre of Competence in Research (NCCR) Robotics through the Swiss National Science Foundation and the SNSF-ERC Starting Grant.
The Titan Xp used for this research was donated by the NVIDIA Corporation.
Konstantinos G. Derpanis is supported by a Canadian NSERC Discovery grant.
He contributed to this work in his personal capacity as an Associate Professor at Ryerson University.}

\clearpage

{\small
\bibliographystyle{ieee}
\bibliography{all}
}

\clearpage

\section{Supplementary Material}

\subsection{CSV files}

As written in the paper, a detailed CSV file is attached for each of the two robotic testing datasets.
This file contains the columns:
\begin{itemize}
  \setlength\itemsep{-.5em}
  \item {\tt name}: sequence name and the indices of the two images that form the pair.
  \item {\tt dR, dt}: ground truth difference in rotation and translation between the two image frames.
  \item {\tt nmin}: number of points that needs to be extracted to obtain $10$ inliers.
  \item {\tt eR, et}: rotation and translation \emph{error} of the relative pose estimate.
\end{itemize}
Both {\tt dR} and {\tt eR} are measured with the geodesic distance, in degrees, which corresponds to the angle of the angle-axis representation of the relative rotation.

\subsection{True pose difference plots}

\begin{figure}
  \centering
  \includegraphics[width=\columnwidth]{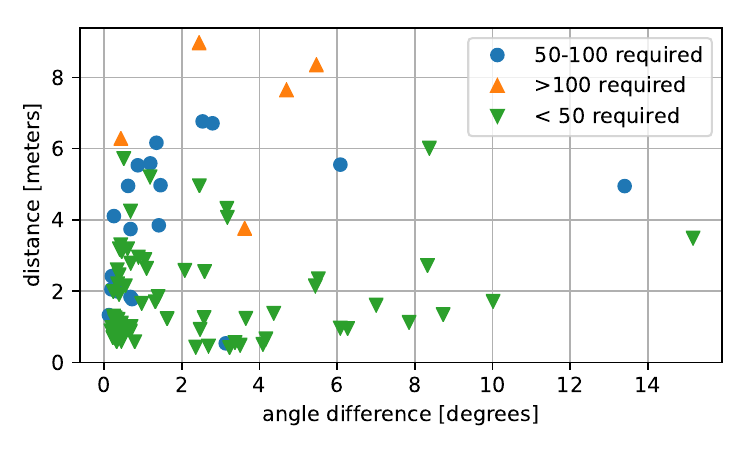}
  \includegraphics[width=\columnwidth]{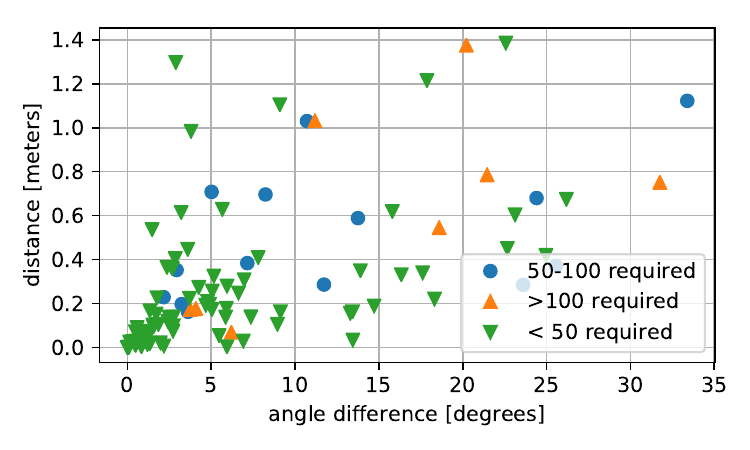}
  \caption{True pose difference plots for KITTI (top) and EuRoC (bottom).
  Different markers indicate different numbers of interest points that are required using our method to obtain $10$ inliers.} 
  \label{fig:dr_dt}
\end{figure}

To provide an intuition for the robotic testing sets, we plot the distribution of true relative poses in \reffig{fig:dr_dt}. 
Note the difference between the two datasets:
KITTI exhibits larger translation distances, while EuRoC exhibits larger orientation differences.
Recall that given the first image, the second image is randomly sampled among subsequent images with a scene overlap of at least $50\%$.

Besides showing the distribution, this plot also indicates how many interest points need to be extracted to obtain $10$ inliers for each pair.
While generally, more points are required at larger pose differences, there are no very clear boundaries --- relative pose is not a strong predictor of succinctness for an image pair.

\subsection{Match renderings}

Several match renderings akin to Figure 1 in the paper are attached.
They show the input image, score output and the inliers when detecting $50$ interest points. 
Probability prediction is indicated with the thickness of the circle, and inlier matches are connected with lines.
The plots demonstrate the peakedness of our response and validates that our method works as expected.

\subsection{Sequence videos}

We visualize the stability of our interest points as well as the correlation of that stability with predicted inlierness probability with a video composed of the score maps extracted over the full testing sequences.

To highlight succinctness, only $50$ points are extracted in each frame.
Note that the interest points are independently extracted in each frame.
Still, many of the interest points behave as if they were directly tracked, especially the ones with high scores, as indicated by thicker circles.
This corroborates our results about succinctness --- few points are needed to obtain sufficient inliers.
Furthermore, the natural peakedness of our score response can be easily discerned in the videos.

\end{document}